\documentclass[runningheads]{llncs}
\usepackage[utf8]{inputenc}

\usepackage{graphicx}
\usepackage{subcaption}
\usepackage{algorithm}
\usepackage{hyperref}
\usepackage[noend]{algpseudocode}
\usepackage{amsmath,amssymb,amsfonts}
\usepackage{wrapfig}

\begin{document}
\title{Reinforcement Learning in a Physics-Inspired Semi-Markov Environment}
%
%

\author{Colin Bellinger\inst{1} \and
Rory Coles\inst{2} \and
Mark Crowley\inst{3} \and 
Isaac Tamblyn\inst{1,4}
}


%

\institute{National Research Council of Canada, Ottawa, Canada\\
\email{\{Colin.Bellinger, Isaac.Tamblyn\}@nrc-cnrc.gc.ca}\\ \and
University of Victoria, Victoria, Canada\\ 
\email{rfcoles@uvic.ca} \and
University of Waterloo, Waterloo Canada \\
\email{mark.crowley@uwaterloo.ca} \and
Vector Institute for Artificial Intelligence, Toronto, Canada}

\maketitle              
\begin{abstract}

Reinforcement learning (RL) has been demonstrated to have great potential in many applications of scientific discovery and design. Recent work includes, for example, the design of new structures and compositions of molecules for therapeutic drugs. Much of the existing work related to the application of RL to scientific domains, however, assumes that the available state representation obeys the Markov property. For reasons associated with time, cost, sensor accuracy, and gaps in scientific knowledge, many scientific design and discovery problems do not satisfy the Markov property. Thus, something other than a Markov decision process (MDP) should be used to plan / find the optimal policy. In this paper, we present a physics-inspired semi-Markov RL environment, namely the phase change environment. In addition, we evaluate the performance of value-based RL algorithms for both MDPs and partially observable MDPs (POMDPs) on the proposed environment. Our results demonstrate deep recurrent Q-networks (DRQN) significantly outperform deep Q-networks (DQN), and that DRQNs benefit from training with hindsight experience replay. Implications for the use of semi-Markovian RL and POMDPs for scientific laboratories are also discussed.

\end{abstract}

\section{Introduction}

Developing new materials is seen as a key to advance in many areas of science and society \cite{Decadal_2019}. Currently, state-of-the-art methods for developing new materials are slow, unpredictable, and have high associated costs. Artificial intelligence has the potential to make significant contributions to problems of this nature.

In recent years, deep reinforcement learning (RL) has achieved significant advancements, and produced human level performance on challenging video games, board games, and in robotics \cite{mnih2013playing,silver2016mastering,gu2017deep}. These results have garnered much attention across a wide variety of domains, including the fields of chemistry and physics. RL has, for example, been applied in quantum physics and chemistry \cite{andreasson2019quantum,zhou2019optimization}. The latter is partially motivated by work with scientific laboratory robots \cite{macleod2019self,roch2018chemos}

Our research focuses broadly on the application of RL to materials science. We hypothesise that RL has a great potential to speed up the materials design and discovery process. From an AI perspective, this application area embodies many interesting challenges. In materials, for example, evaluating prospective solutions can be costly, time consuming and destructive. Therefore, sample efficiency is a key requirement. On the other hand, an agent may have multiple goals, and/or new goals may be added overtime. Thus, multi-agent learning with shared experience and transfer learning are of interest. The rewards are often binary and significantly delayed, which motivates the need for strategies to handle rewards, and improve sample efficiency. Moreover, important information to the materials design process is often hidden due to costs and scientific limitations. Thus, the AI must be suitable for semi-Markov decision processes.

To date, there has not be a systematic investigation of the suitability of deep RL algorithms for applications in materials science involving semi-Markov decision processes. In this paper, we commence this exploration by presenting a new physics-inspired semi-Markov learn task; specifically the semi-Markov phase change environment. Subsequently, we conduct an initial evaluation of the potential for value-based deep RL algorithms in the environment, and discuss the challenges to be faced in future real-world applications. 

\subsection{Contributions}

We make the following contributions in this paper:
\begin{itemize}
    \item Introduce the semi-Markov phase change environment;
    \item Compare the performance of deep Q-networks (DQN) to deep recurrent Q-networks (DRQN) on the proposed environment;
    \item Evaluate the benefit of hindsight experience replay (HER) on DQN and DRQN; and,
    \item Discuss the performance gap between these methods and the optimal policy.
\end{itemize}

\section{Related Work}

Q-learning is an off-policy temporal difference control algorithm \cite{watkins1989learning} where the objective is to learn an optimal action-value function, independent of the policy being followed. DQN is a recent variation of Q-learning that takes advantage of the generalizing capabilities of deep learning. DQNs have been shown to produce human-level performance on challenging games on Atari 2600 \cite{mnih2015human}.

DQNs offer a solution approach for Markov decision processes (MDPs). Specifically, problems where the state observation emitted from the environment is sufficient to select the next action. Cases where the Markov property does not hold, require a \textit{partially observable MDP (POMDP)}. In these cases, the representation of the current state alone is not sufficient to select the next action. This can occur due to unreliable observations, an incomplete model (i.e. latent variables), noisy state information or other reasons. 

In \cite{hausknecht2015deep}, the authors propose the use of a recurrent neural network architecture in place of the feed-forward network in DQN. Leveraging recurrent neural networks, it is argued, enables the Q-network to better handle POMDPs. Specifically, with the recurrent neural network, the agent can build an implicit notion of its current state based on the recent sequence of state observation resulting from actions taken. The authors show that Deep Recurrent Q-Networks (DRQN) presented with a single frame at each time-step can successfully integrate information through time, and thereby replicate the performance of DQNs on standard Atari 2600 games. In this work, we extend the evaluation of DRQNs to the phase change environment in order to better understand the potential of DRQN on real-world POMDPs. 

Many of the recent achievements of deep RL have been produced in simulated environments because RL agents must gather a large amount of experience. Deep Q-Networks, for example, famously required approximately 200 million frames of experience for traininga and  approximately 39 days of real-time game playing, on the Atari 2600 \cite{mnih2015human}. Model-based RL methods, such as DYNA-Q \cite{sutton1990integrated}, aim to replace a large portion of the agent's real-world experience with experience collected from a surrogate or other models of the environment. Model-based methods, however, have seen most of their successes in environments where the dynamics are simple and can easily and accurately be learned. This is decidedly \textit{not} the case for most physics and chemistry environments.

Learning in many physics and chemistry environments is made more challenging by sparse, binary rewards. Andrychowicz \textit{et al.,} in \cite{andrychowicz2017hindsight}, proposed Hindsight Experience Replay (HER), which extends the idea of training a universal policy \cite{schaul2015universal}. Inspired by the benefit that humans garner by learning from their mistakes, HER simulates this by re-framing a small, user-defined, portion of the failed trajectories as successes. It is applicable to off-policy, model-free RL, and to domains in which multiple goals could be achieved. HER was shown to improve sample efficiency, and make learning possible in environments with sparse and binary reward signals. 

Multi-goal learning environments with sparse, binary rewards, and the necessity for sample efficiency are key features of many physics and chemistry applications, such as materials design. As a result, HER is potentially of great value in these domains. To date, however, it has not been evaluated in semi-Markov decision processes nor has it been explored in conjunction with DRQN. 

\section{Semi-Markov Phase Change Environment}

\begin{figure}
\centering
\includegraphics[scale=0.45]{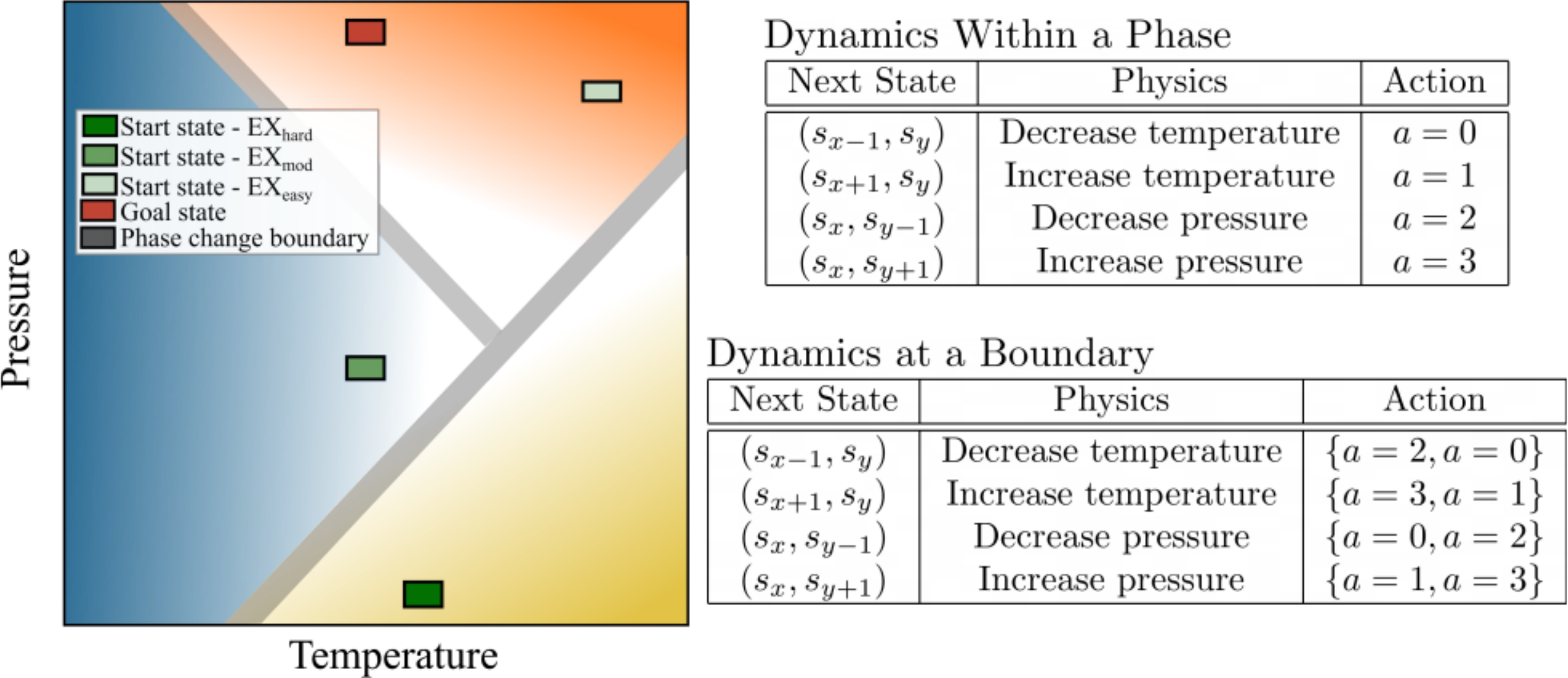} 
\caption{(Left) Phase change environment. (Top right) State transition dynamics for within-phase states. (Bottom right) State transition dynamics for phase change boundary. This assumes that the agent is at some boundary states.} \label{fig:phaseChangeEnv}
\end{figure}

Our new semi-Markov phase change environment\footnote{The environment is  is available at \url{http://clean.energyscience.ca/gyms}.} is implemented based on the OpenAI Gym framework \cite{openAIGym2016} and is depicted in Figure \ref{fig:phaseChangeEnv}. Within the physical sciences, Figure \ref{fig:phaseChangeEnv} is known as a phase diagram - a convenient representation of a materials behaviour where, within a ``phase'', symmetry is preserved over a wide range of experimental conditions (in this case temperature, $T$, and pressure, $P$). In general, it is possible to alter the pressure or temperature of a material while remaining within the same phase (e.g. cold water and warm water are both liquids). Within a single phase, adding heat ($Q^+$) results in a positive change of temperature, while removing it, ($Q^-$) does the opposite. Similarly, within a single phase, doing positive work ($W^+$) increases the pressure, while negative work, ($W^-$) results in a pressure decrease. 

Importantly, we note that the relationship between heat, work, temperature, and pressure is different at the boundary between some phases. Thus, the state transition dynamics are different at the boundary. Specifically. symmetries change when crossing a discontinuous phase boundary (e.g solid-liquid). This change is accompanied by the addition or removal of a latent heat. On a phase diagram, such a boundary is denoted with a solid line. Because of the latent heat, under equilibrium conditions, two or more phases can co-exist with one another in a stable state. As a result, when visualized on a phase diagram, a trajectory of constant heating will temporarily stall at a phase boundary. There is an apparent lack of progress at the boundary while this energy is used to convert the material from one phase into another at constant $(P,T)$ (e.g. the size of an ice-cube decreases while the amount of liquid water increases).

In our environment, the agent's goal is to take a series of actions that add or remove energy in the system by two independent mechanisms (heat and work) to modify a material from its start state $M_s$ to some goal state $M_g$. The result of the actions is measured in terms of the pressure and temperature of the material. 

The environment has a discrete $4$-action space, $\mathcal{A} \in \{a_0=Q^-,a_1=Q^+,a_2=W^-,a_3=W^+\}$. Thus, the agent must learn to navigate from some start position in the two-dimensional temperature-pressure space $M_s = (t_s,p_s)$ to some goal state $M_g = (t_g,p_g)$ in as few steps as possible. The episodes terminates immediately after the agent takes the action to transition in to $M_g$. The agent receives a reward of 1 when it reaches the goal, and zero elsewhere. The optimal policy in the environment is to apply the minimum number of actions (steps) to get to the goal. The environment emits a state observations in terms of $T$ and $P$. The initial version of the environment has discretized pressure and temperature measurements, and a limit to the range. This results in a $2$-dimensional grid state-space with vertical movements analogous to changes in pressure (resulting from $W^+/-$) and horizontal movements corresponding to changes in temperature (resulting from the $Q^+/-$).

The environment is designed to weakly approximate the process of adding small, fixed amounts of energy (in the form of heat or work) to an initial phase (e.g. a liquid) to convert it to another one (or for the case of the phase boundary, a mixture of different fractions solid, liquid, and gas). In order to make the problem extra challenging, we include the requirement that the agent invoke two different actions when it crosses through the boundary. While this would not strictly be required physically for equilibrium processes, it makes the learning task more difficult and relevant for real world examples which involve nucleation, activation barriers, etc.

Unlike the traditional grid-world setup, the grid-based phase change environment does not have any barriers that might prevent an agent from moving in a certain direction. The challenge, as we have discussed from the scientific perspective above, is learning to efficiently navigate through partially observable phase change boundaries. The state transition dynamics are presented in the tables on the right in Figure \ref{fig:phaseChangeEnv}.  

The dynamics for the phase change boundary are as such, when in some boundary state $(s_x, s_y)$, the agent must apply a sequence of two actions to transition into the state on the other side of the boundary. In order for the agent to move in the direction of increasing pressure, for example, it must apply action $a=1$ followed by action $a=3$. This leads to the following state-action sequence:

\begin{equation}
    ...(s_x, s_y), a=1, (s_x, s_y), a=3, (s_x, s_{y+1})...
\end{equation}
\noindent

\section{Experimental Setup}

\subsection{Reinforcement Learning Algorithms}

In order to assess the suitability of value-based RL in a semi-Markov materials-inspired environment, we compare the performance of DQN to DRQN. We evaluate DRQN with a trace length of one (\textit{i.e.}, a one state history) as this makes it directly comparable to DQN. This forces the network to rely solely on its internal architecture to remember the implicit state of the system. Finally, we explore the benefit of HER on DQN and DRQN in the semi-Markov environment.  

\subsubsection{Deep Q-Learning: } In this work, DQN receives a state vector $s=[P,T]$ as the input and emits a value for each action $Q = (a_1, a_2, a_3, a_4)$ at the output layer. A greedy agent in state $s$ will take $a= \text{arg\_max} ~Q$. The parameters of the network $\theta$ are updated as, $\theta_{i+1} = \theta_i + \alpha\Delta_\theta \mathcal{L}(\theta_i)$, to minimize the loss function, $\mathcal{L}(s,a|\theta) \approx (r + \gamma \max_a \mathcal{Q}(s^\prime, a|\theta) - \mathcal{Q}(s,a|\theta))^2$, where $r$ is the reward, $\gamma$ is a discount factor and $\alpha$ is the learning rate. In this case, a single network is generating the update target and being updated. Updating based on a single network has been shown to lead to instability in some cases, and can be improved upon by having a separate target network. However, this was not necessary in the phase change environment.

In the following experiments, we applied a neural network with a single 48 unit hidden layer with ReLU activation and the ADAM optimizer. The free parameters were set as follows, the discount factor $\gamma=0.95$ and the exploration rate $\epsilon=1$ with linear $\epsilon_\delta = 0.00001$ decay. After an initial period of experience gathering, the network was updated after every episode by sampling a batch of size $127$ from the experience replay buffer.

\subsubsection{Deep Recurrent Q-Learning: }

DRQN follows the same setup as that presented for DQN above. Specifically, the input, output and objective function, and optimizers are the same. The key difference is that the fully connected hidden layer in DQN is replaced by a recurrent network. In our experiments below, we use a 128 unit Gated Recurrent Unit. 

\subsubsection{Training with Hindsight Experience Replay: }

HER is a training framework that requires the current state and the goal state to jointly form the state space. Thus, all experiments related to HER have an expanded state space. We edited $5\%$ of the tuples corresponding to failed actions (\textit{i.e.}, action with zero reward) to be seen as successful. Specifically, we set the reward to $1$ and the goal state to the current state, prior to adding the tuple to the experience replay buffer. 

\subsection{Evaluation Method}

In order to thoroughly assess the impact of the non-Markov phase change boundaries on the RL algorithms, we evaluate each method from three deterministic starting locations. From each of these starting locations, the agents must learn to navigate to a single goal. In experiment 1 (EX$_\text{hard}$), the agents start off farthest from the goal and must cross two phase change boundaries. The agent starts marginally closer to the goal in experiment 2 (EX$_\text{mod}$). Here, the agent must cross a single non-Markov barrier. Finally, in experiment 3 (EX$_\text{easy}$) the agent starts close to the goal and is not required to cross any non-Markov barriers. 

To further our analysis of the impact of the non-Markov phase change boundaries on the RL algorithms, we repeat each of the above experiments in a Markov version of the phase change environment. In the Markov version, the dynamics in the phase change boundaries are equivalent to the inner-phase dynamics, and all of the states are fully observable. 

The performance of each agent is recorded on intervals of $50$ episodes. Specifically, after each increment of $50$ episodes of training, each agent is applied for one episode (or a maximum of $10,000$ steps) of testing with an $\epsilon$-greedy policy ($\epsilon=0.2$). Thus, for $20,000$ episodes of training, in each of the $30$ iterations, we collect 400 test results. These are averaged and reported in the plots below.




\section{Results}

\subsection{DQN on the semi-Markov phase change environment}

\begin{figure}
\centering
\includegraphics[scale=0.35]{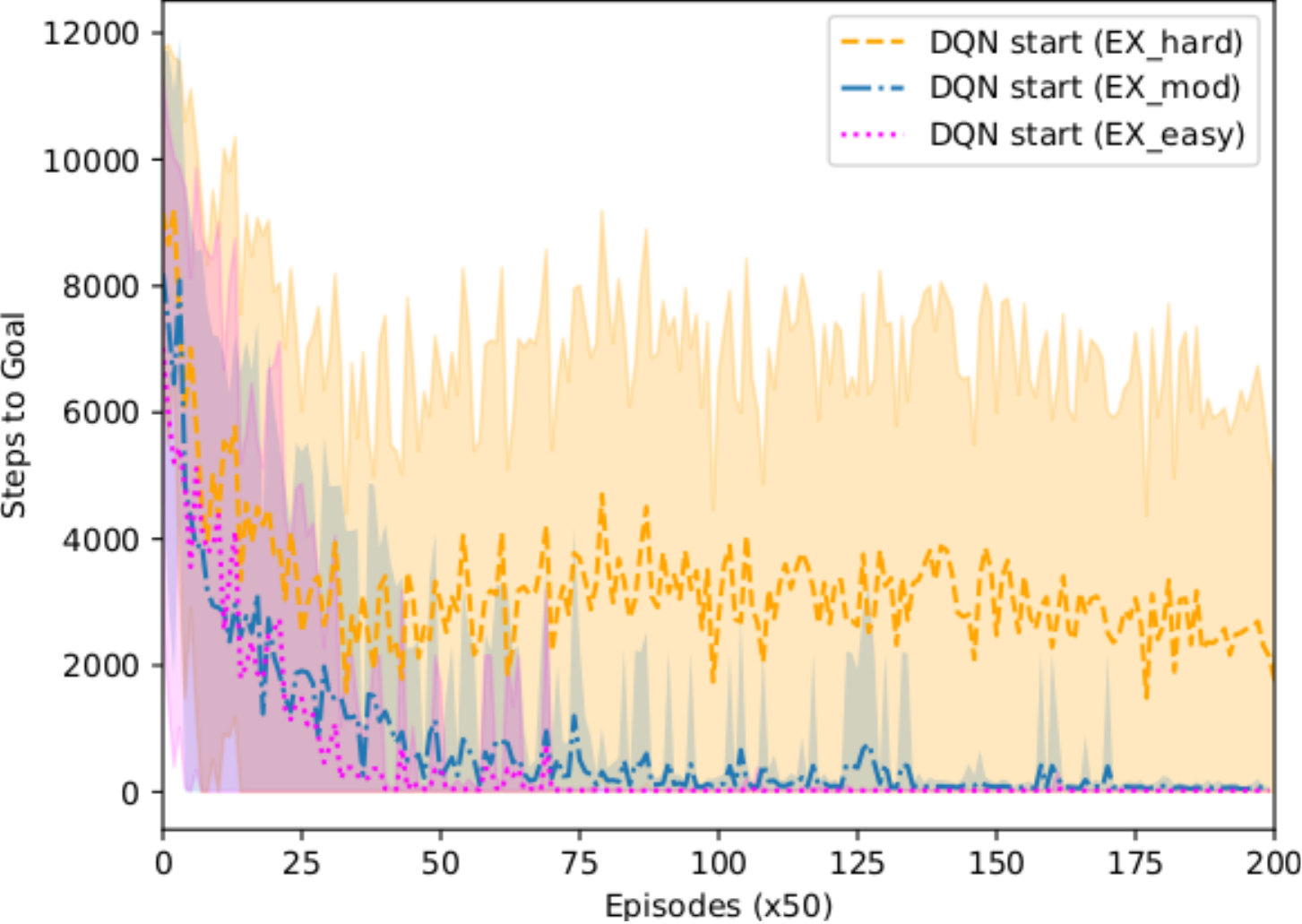}
\includegraphics[scale=0.35]{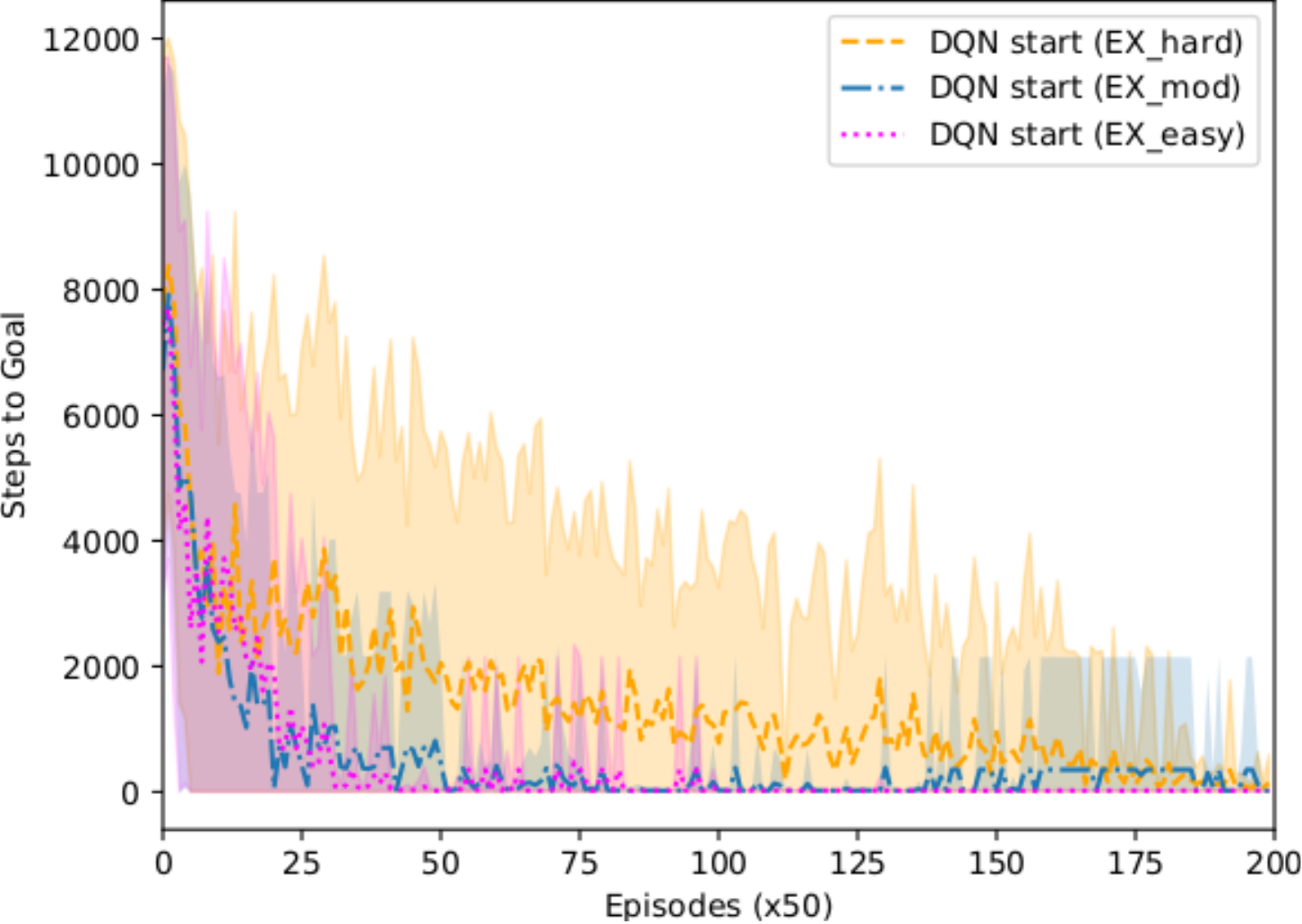}
\caption{Mean number of steps per episode for DQN on (left)  the semi-Markov phase change environment, and (Right) the Markov phase change environment} 
\label{fig:dqnPCE}
\end{figure}

The plot on the left in Figure \ref{fig:dqnPCE} shows the average number of steps per episode that an agent learning with DQN takes to the goal in the semi-Markov environment when starting at each of the three starting locations EX$_\text{easy}$,EX$_\text{mod}$, and EX$_\text{hard}$. For comparison, the results on the right show the performance when the phase change environment is made fully Markovian.

These results demonstrate that DQN is affected by both the distance between the starting state and goal (sparsity of reward) and semi-Markov decision process resulting from the phase change boundaries. From the left plot, it is clear that the agent in EX$_\text{hard}$ learns much slower than the agents in EX$_\text{mod}$ and EX$_\text{easy}$. The mean number of steps by episode for EX$_\text{mod}$ and EX$_\text{easy}$ are nearly indistinguishable, whereas the mean number of steps for EX$_\text{hard}$ remains significantly higher throughout training. Two factors are contributing to this, the crossing of phase change boundaries and the distance from the goal state. 

To understand which factor is impacting the performance in EX$_\text{hard}$ more, we compare the corresponding plots on the left (semi-Markov) and the right (Markov). On the semi-Markov environment, initially the mean number of steps drops quickly, before plateauing at what is still a large mean number of steps to the goal. Alternatively, in the Markov environment, the agent starting from EX$_\text{hard}$ consistently learns to take fewer steps to the goal. Here, it converges to a mean number of steps that is much closer to optimal. This suggests that while DQN is harmed by the reward sparsity, it is the non-Markov phase change boundaries that prevent it from converging to the optimal number of steps.

\subsection{DRQN on the semi-Markov phase change environment}

Given the significant effect caused by the non-Markov phase change boundaries, we now investigate the extent to which the hidden representation and sequential nature of recurrent neural networks enables agents learning with DRQN to better navigate the non-Markov phase change boundaries. 

\begin{figure}
\centering
\includegraphics[scale=0.35]{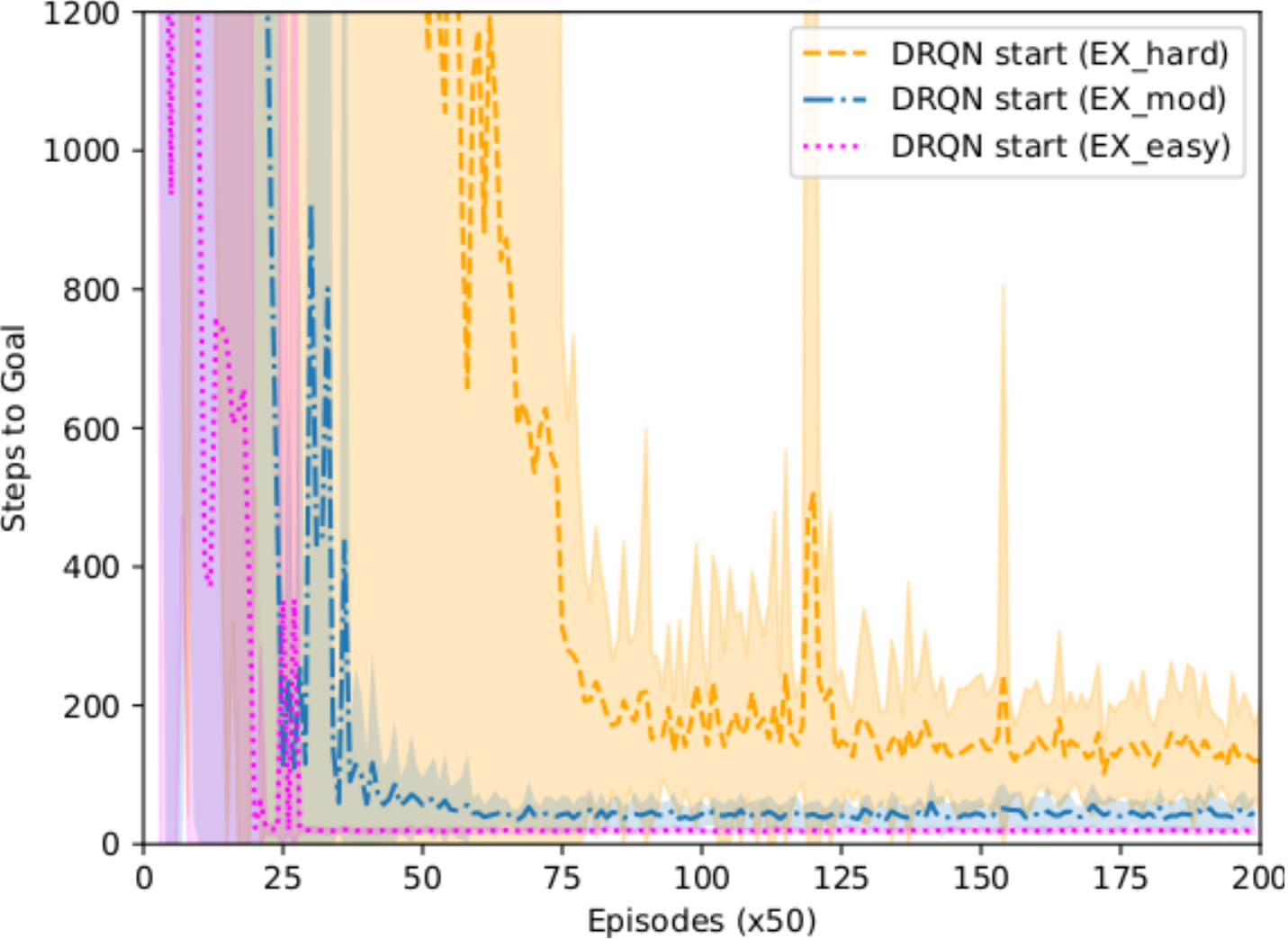}
\includegraphics[scale=0.35]{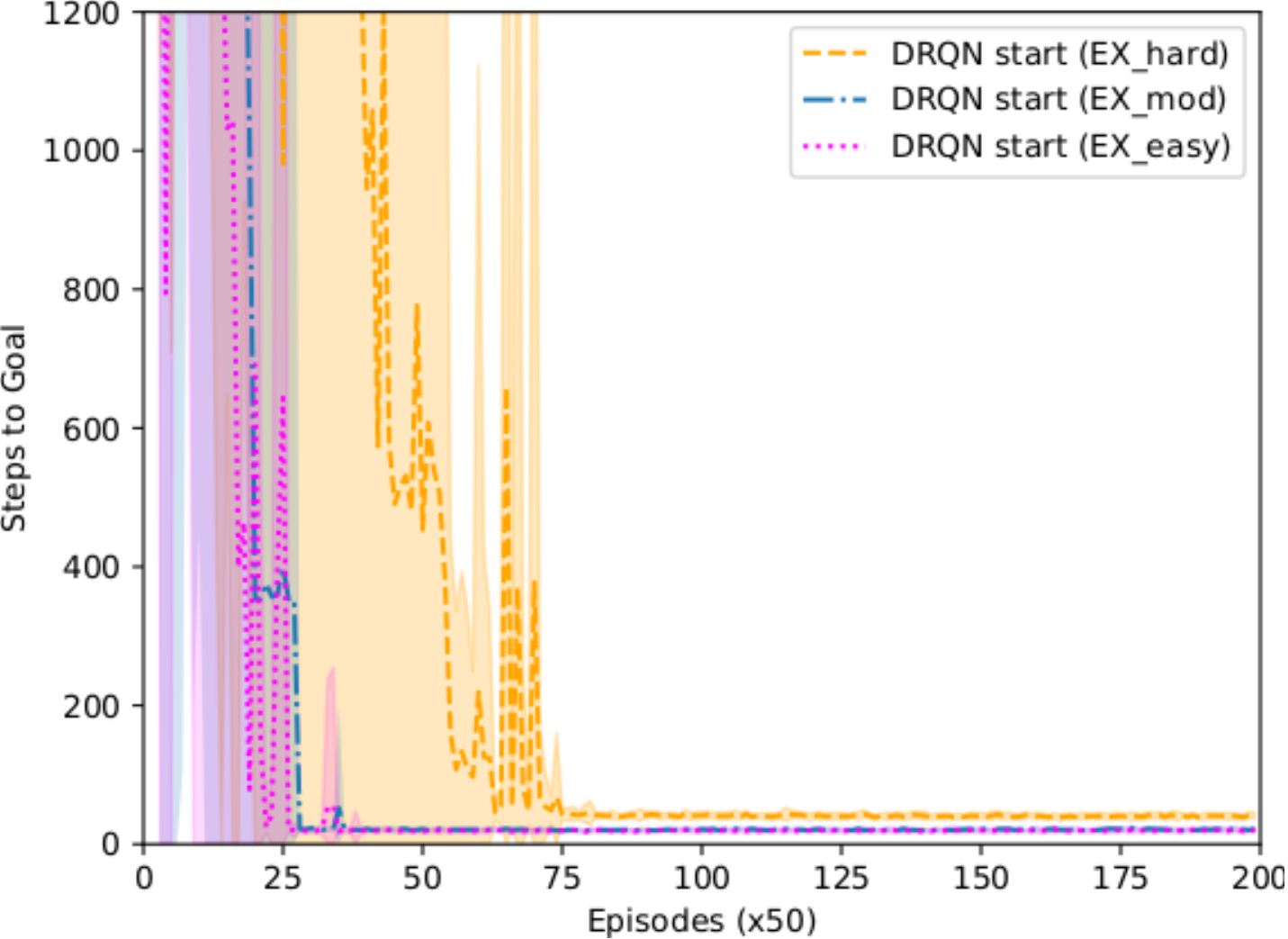}
\caption{(left) Mean number of steps per episode for DRQN on the semi-Markov phase change environment. (Right) Mean number of steps per episode for DRQN on the Markov phase change environment} \label{fig:drqnPCE}
\end{figure}

The left plot in Figure \ref{fig:drqnPCE} shows the average number of steps per episode that the agent learning with DRQN takes on route to the goal in the semi-Markov environment. The results for the Markov version of the phase change environment are shown on the right.

For the most challenging case EX$_\text{hard}$, DRQN converges to approximately $530$ steps after $10,000$ episodes ($200 \times 50$.) By contrast, when the agent learns with DQN on EX$_\text{hard}$, it does not converge after $20,000$ episodes of training. Thus, DRQN provides a good improvement in terms of the convergence speed and the average number of steps taken on route to the goal. 

Comparing the semi-Markov results on the left and the Markov results on the right reveals that the DRQN agent on the semi-Markov problem is still not equivalent to the agent on the Markov problem. The gap, however, is closed significantly from what we found with DQN. In the Markov environment, the DRQN agent in EX$_\text{hard}$ converges after approximately $3,750$ episodes to $40$ steps (which is optimal), versus approximately $530$ steps after around $10,000$ episodes for the semi-Markov environment. 

\subsection{Agents With Hindsight Experience Replay}

The above results demonstrate that learning with DRQN can produce a significant reduction in the number of steps taken to the goal, and a significant speed up in the rate of learning in comparison to DQN. Nonetheless, the number of episodes DRQN requires to converge is more than double on the semi-Markov problem, and the converged agent takes on average over $10$ times more steps. 

In the following two subsections, we evaluate whether HER helps to improve the rate of convergence on the semi-Markov phase change environment, and assess how it compares to the Markov environment.

\subsubsection{DQN + HER:}

\begin{figure}
\centering
\includegraphics[scale=0.35]{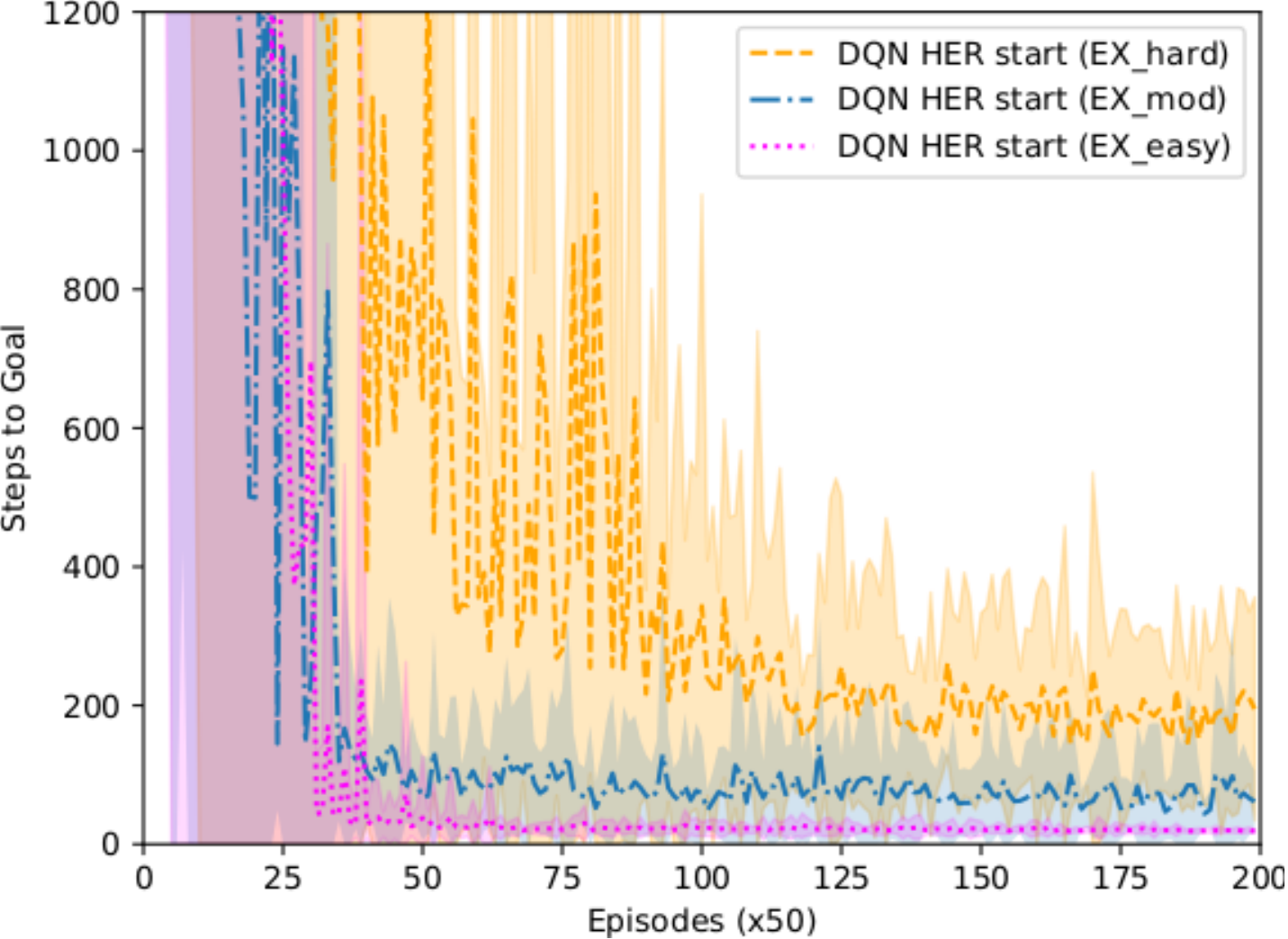}
\includegraphics[scale=0.35]{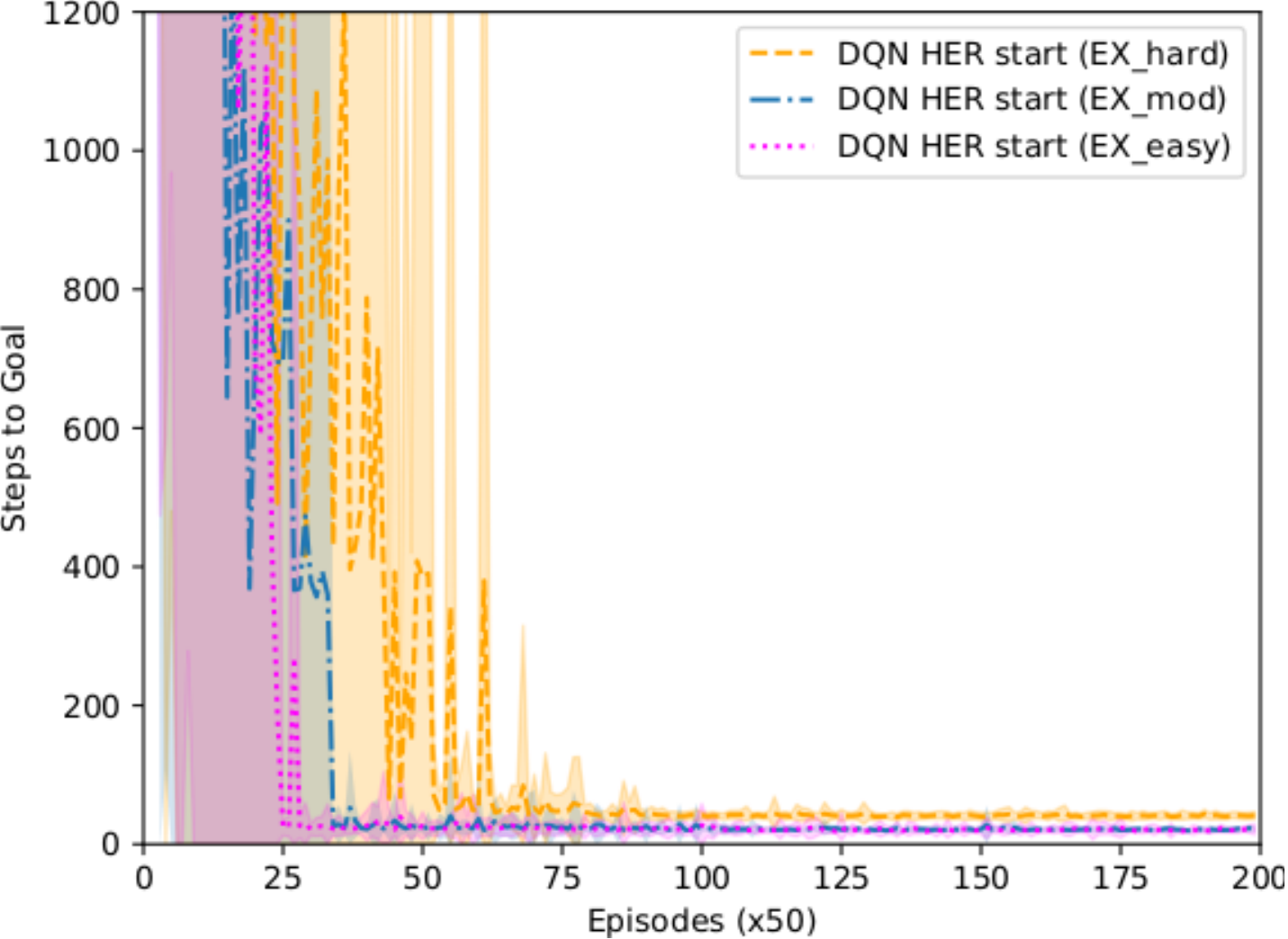}
\caption{(left) Mean number of steps per episode for DQN with HER on the semi-Markov phase change environment. (Right) Mean number of steps per episode for DQN with HER on the Markov phase change environment} \label{fig:dqnHerPCE}
\end{figure}

Figure \ref{fig:dqnHerPCE} shows the mean performance for DQN with HER in the semi-Markov phase change environment on the left and in the Markov environment on the right. Once again, we will focus on the performance on EX$_\text{hard}$ as it produces the most insightful results. The plot demonstrates that the DQN agent learns significantly faster with HER than without. This is consistent with previously published results. In particular, the agent on EX$_\text{hard}$ converges to approximately $195$ steps after $6,000$ episodes of learning on average. Whereas, without HER, DQN had not converged after $20,000$ episodes. Interestingly, this is faster convergence, and to fewer steps than the DRQN results reported in the previous section. This is likely due to improved efficiency within the phases, whilst the accuracy of the action selection in the non-Markov phase change boundaries remains less than optimal.  The performance gap with the Markov environment is narrowed, but still wide.  Specifically, in the Markov setup, DQN + HER converges to approximately $43$ steps (approximately optimal) after an average of $3,150$ episodes. 

\subsubsection{DRQN + HER:}

\begin{figure}
\centering
\includegraphics[scale=0.35]{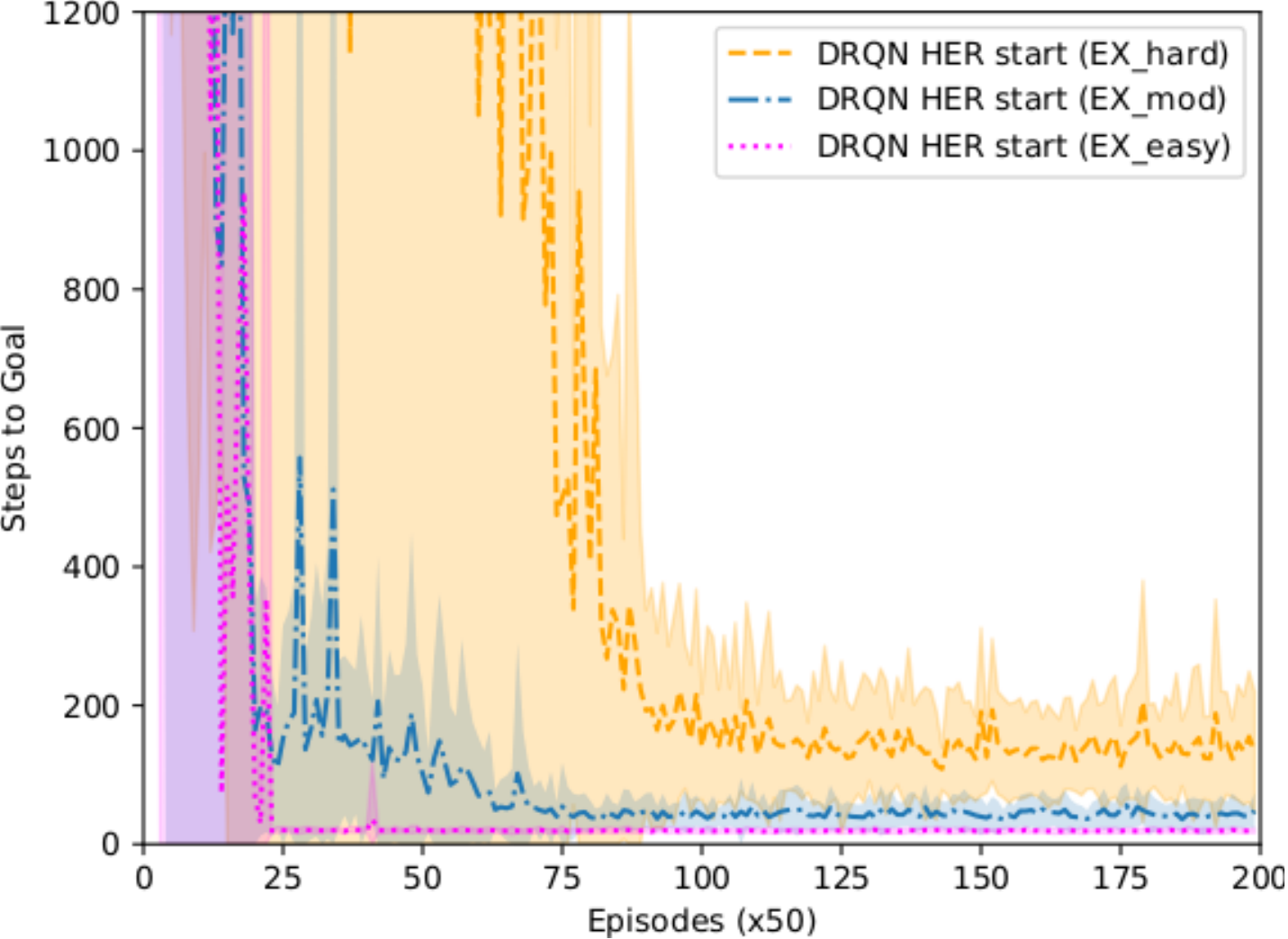}
\includegraphics[scale=0.35]{dqnHerCompareStartLocationsSub.pdf}
\caption{(Left) Mean number of steps per episode for DRQN with HER on the semi-Markov phase change environment. (Right) Mean number of steps per episode for DQN with HER on the semi-Markov phase change environment (re-plotted for ease of analysis).} \label{fig:drqnHerPCE}
\end{figure}

Finally, we evaluate the benefit of using HER with DRQN in the semi-Markov phase change environment. These results are presented in Figure \ref{fig:drqnHerPCE}. The earlier results with DRQN on  EX$_\text{hard}$ amounted to $530$ steps after approximately $10,000$ episodes. With the addition of HER, the agent converges to approximately $145$ steps on average after $4,000$ episodes. This shows that DRQN receives a good performance boost from the addition of HER in terms of average number of steps and the rate convergence. For comparison, the agent learning with DQN + HER on  EX$_\text{hard}$ converges to approximately $195$ steps after $6,000$ episodes of learning on average. Thus, DRQN + HER is the better of the two methods on the semi-Markov phase change environment. 

Despite its superiority, there is a noteworthy lag in the learning curve for DRQN + HER for EX$_\text{hard}$ before the mean number of steps steeply drop off. Whereas, DQN + HER has a relatively consistent drop in the mean number of steps from the outset. This difference suggests that agents learning with recurrent neural network models may suffer from an initial lag in performance due to the added complexity of training the GRU. 

\section{Discussion}

Our results have extended the previous analysis of DRQN as a method to solve POMDPs to problems beyond the standard Atari 2600 game suite. In particular, our results show that agents trained with DRQN learn significantly better value-functions for a physics-inspired semi-Markov phase change environment in comparison to DQN. Specifically, adding the recurrent architecture to the DQN enables the agent to takes fewer steps on route to the goal. Moreover, we show that DRQN is further improved in terms of the learning rate and the number of steps to the goal when HER is incorporated into the training process.

In spite of the significantly improved performance, DRQN does not learn a value-function that implements a perfect policy for the semi-Markov phase change environment. After convergence, DRQN + HER takes on average $3$-times the optimal number of steps on route to the goal in EX\_hard in the semi-Markov environment. Without HER, DRQN takes on average 13-times more steps than optimal. As can be seen in Figure \ref{fig:drqnHerPCE}, the gap is significantly narrowed for EX\_mod, and is completely closed for EX\_easy. This suggests that the portion of non-Markov states has a non-linear impact of the learning difficulty. 

A potential method to improve the performance of the DRQN is to use longer trace length. Longer trace lengths would provide more direct information about the state sequence, and potentially simplify the problem. Our current analysis does not reveal where the extra steps are taken. Nonetheless, it highly likely that the agent would still struggle with the semi-Markov phase change boundaries. Our ongoing research aims to identify where the DRQN is failing to learn the optimal actions in order to propose improvements. There is clearly room for improvement; here we have established a strong baseline for future work.

From an experimental science perspective, these results suggest that RL has the potential to have a significant, positive impact of the advancement of materials, and other experimental science. We note, however, that application of RL in the laboratories will involve several more layers of complexities on top of partial observability. Each of these challenges needs to be clearly understood and analyzed from an RL perspective in order to leverage the right tools from the current state-of-the-art and to develop new RL theories and methodologies where necessary. In the list below, we outline a few characteristic of laboratory learning that we see as being pertinent.
\begin{itemize}
\item The existence of different classes of sensors, each of which provide different information content, costs, and data representation;
\item The value and cost of each sensor depends on time and space;
\item Because observations are costly, the agent should have the ability to make active decisions about when to take an observations and which observations to make;
\item Sensors have intrinsic quantifiable uncertainties associated with them.
\item In a significant number of experiments there is a simple phenomenological model which can roughly predict the outcome. 
\end{itemize}

A straightforward extension of the results presented here would be to include simulated spectroscopic sensor input. This is closer to the conditions that human operators face. Additionally, in our simple model of material phases, the mapping between energy input to change in conditions (P,T) did not vary across the different phases. In general, this is not true and depends on the specific heat and compressibility of the material. Finally, throughout, we assumed equilibrium conditions - i.e. the timescale of internal relaxations was short compared to the observation time. 

%



\section{Conclusion}

We introduced the phase change environment to evaluate RL algorithms on a semi-Markov problem inspired by physics and laboratory science. We compared DQN and DRQN with and without HER in the environment. Our results show that DRQN learns significantly faster and converges to a better solution than DQN in this domain. Moreover, we find that the number of episodes to convergence in DRQN is further improved by the incorporation of HER. Nonetheless, the hypothesis that the implicit state estimate maintained by the recurrent network in DRQN would enable it to learn to behave optimally in the phase change environment was not realized in these experiments. Specifically, DRQN+HER converges to approximately 3-times the optimal number of steps on EX$_{hard}$.  

Our ongoing research is evaluating the benefit of longer trace lengths for DRQN and alternative algorithms for semi-Markov decisions processes. In addition, we are developing more materials-inspired RL environments to evaluate existing algorithms and promote the development of new, superior algorithms for materials design and discovery.

\section{Acknowledgements}

Work at NRC was performed under the auspices of the AI4D Program.

\bibliographystyle{plain}
\bibliography{library}

\end{document}